\documentclass[a4paper,onecolumn,10pt,times]{article}
 
\usepackage[dvips]{graphicx}
\usepackage{latexsym}
\usepackage{amsmath} 
\usepackage{amssymb} 
\usepackage{graphics} 

\makeatletter
\def\@maketitle{%
  \vbox to 6.5cm{%
    \hsize\textwidth
    \linewidth\hsize
    \vspace{1.5cm}
    \centering
    {\bfseries\LARGE \@title \par}
    \vspace{24pt}
    {\fontsize{11pt}{13pt}\selectfont \begin{tabular}[t]{c}\@author \end{tabular}\par}
    \vfill} 
}

  {\small
    \list{}{\labelwidth0pt
      \leftmargin0pt \rightmargin\leftmargin
      \listparindent\parindent \itemindent0pt
      \parsep0pt
      }%
    \item[\hskip\labelsep\bfseries\abstractname\enspace --] \itshape}{\endlist}

\newcommand{\keywordsname}{Keywords}
  {\small
    \list{}{\labelwidth0pt
      \leftmargin0pt \rightmargin\leftmargin
      \listparindent\parindent \itemindent0pt
      \parsep0pt
      }%

    \item[\hskip\labelsep\bfseries\keywordsname:]}{\endlist}

\begin{document}

\title{\bf Uniform and Partially Uniform Redistribution Rules}

\author{
Florentin Smarandache\\
Department of Mathematics\\
University of New Mexico\\
Gallup, NM 87301, U.S.A.\\
smarand@unm.edu\\
\and Jean Dezert\\
ONERA\\
The French Aerospace Lab,\\
F-91761 Palaiseau, France.\\
Jean.Dezert@onera.fr
}
  
\maketitle    
\thispagestyle{empty}

\noindent
{\bf Abstract -
{\small\em This short paper introduces two new fusion rules for combining quantitative basic belief assignments. These rules although very simple have not been proposed in literature so far and could serve as useful alternatives because of  their low computation cost with respect to the recent advanced Proportional Conflict Redistribution rules developed in the DSmT framework.}}

\vspace{0.5cm}


\section{Introduction}

Since the development of DSmT (Dezert-Smarandache Theory) in 2002 \cite{DSmTBook_2004a,DSmTBook_2006}, a new look for information fusion in the framework of belief has been proposed which covers many aspects related to the fusion of uncertain and conflicting beliefs.  Mainly, the fusion of quantitative or qualitative belief functions of highly uncertain and confliction sources of evidence with theoretical advances in belief conditioning rules. The Shafer's milestone book \cite{Shafer_1976} introducing the concept of belief functions and Demspter's rule of combination of beliefs has been the important step towards non probabilistic reasoning approach, aside Zadeh's fuzzy logic \cite{Zadeh_1975,Zadeh_1979}. Since Shafer's seminal work, many alternatives have been proposed to circumvent limitations of Dempster's rule pointed out first by Zadeh in \cite{Zadeh_1979b} (see \cite{Sentz_2002} and \cite{DSmTBook_2006} for a review). The Proportional Conflict Redistribution rule number 5 (PCR5) \cite{DSmTBook_2006} is one of the most efficient alternative to Dempster's rule which can be used both in Dempster-Shafer Theory (DST) as well as in DSmT. The simple idea behind PCR5 is to redistribute every partial conflict only onto propositions which are truly involved in the partial conflict and proportionally to the corresponding belief mass assignment of each source generating this conflict. Although very efficient and appealing, the PCR5 rule suffers of its relative complexity in implementation and in some cases, it is required to use simpler (but less precise) rule of combination which requires only a low complexity. For this purpose, we herein present two new cheap alternatives for combination of basic belief assignments (bba's): the Uniform Redistribution Rule (URR) and the Partially Uniform Redistribution Rule (PURR). In the sequel, we assume the reader familiar with the basics of DSmT, mainly with the definition and notation of hyper-power set $G^\Theta$ and also bba's defined over hyper-power set. Basics of DSmT can be found in chapter 1 of \cite{DSmTBook_2004a} which is freely downloadable on internet.

\section{Uniform Redistribution Rule}

Let's consider a finite and discrete frame of discernment $\Theta$, its hyper-power set $G^\Theta$ (i.e. Dedekind's lattice) and two quantitative basic belief assignments $m_1(.)$ and $m_2(.)$ defined on $G^\Theta$ expressed by two independent sources of evidence.\\

The  Uniform Redistribution Rule (URR) consists in redistributing the total conflicting mass $k_{12}$ to all focal elements of $G^\Theta$ generated by the consensus operator. This way of redistributing mass is very  simple and URR is different from Dempster's rule of combination \cite{Shafer_1976}, because Dempster's rule redistributes the total conflict proportionally with respect to the masses resulted from the conjunctive rule of non-empty sets. PCR5 and PCR4 \cite{DSmTBook_2006} do proportional redistributions of partial conflicting masses to the sets involved in the conflict. Here it is the URR formula for two sources: $\forall A\neq \emptyset$, one has

\begin{equation}
m_{12URR}(A)=m_{12}(A)+\frac{1}{n_{12}}\sum_{\substack{X_1,X_2\in G^\Theta \\ X_1\cap X_2=\emptyset}} m_1(X_1)m_2(X_2)
\label{eq:URR12}
\end{equation}

\noindent
where $m_{12}(A)$ is the result of the conjunctive rule applied to belief assignments $m_1(.)$ and $m_2(.)$, and $n_{12}=Card\{Z\in G^\Theta, m_1(Z)\neq 0 \ \text{or} \ m_2(Z)\neq 0\}$.\\

\noindent For $s\geq 2$ sources to combine: $\forall A\neq \emptyset$, one has
\begin{equation}
m_{12\ldots sURR}(A)=m_{12\ldots s}(A)
+\frac{1}{n_{12\ldots s}}
\sum_{\substack{X_1,X_2,\ldots ,X_s\in G^\Theta \\ X_1\cap X_2\cap \ldots \cap X_s=\emptyset}} \prod_{i=1}^{s}m_i(X_i)
\label{eq:URR1s}
\end{equation}

\noindent
where $m_{12\ldots s}(A)$ is the result of the conjunctive rule applied to $m_i(.)$, for all $i\in \{ 1,2,\ldots, s\}$ and 
\begin{equation*}
n_{12\ldots s}=Card\{Z\in G^\Theta, m_1(Z)\neq 0 \ \text{or} \ m_2(Z)\neq 0
\ \text{or} \ \ldots \ \text{or} \ m_s(Z)\neq 0\}
\end{equation*}

As alternative, we can also consider the cardinal of the ensemble of sets whose masses resulted from the conjunctive rule are non-null, i.e. the cardinality of the core of conjunctive consensus:

$$n_{12\ldots s}^c=\text{Card}\{Z \in G^\Theta, m_{12\ldots s}(Z)\neq 0\}$$

\noindent We denote this modified version of URR as MURR in the sequel.

\section{Example for URR and MURR}

\label{sec3}

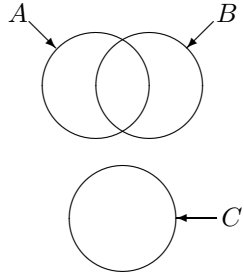
\begin{figure}[!h]
\begin{center}
{\tt \setlength{\unitlength}{1pt}
\begin{picture}(90,90)
\thinlines    
\put(40,60){\circle{40}}
\put(60,60){\circle{40}}
\put(50,10){\circle{40}}
\put(15,84){\vector(1,-1){10}}
\put(7,84){$A$}
\put(84,84){\vector(-1,-1){10}}
\put(85,84){$B$}
\put(85,10){\vector(-1,0){15}}
\put(87,7){$C$}
\end{picture}}
\end{center}
\caption{Hybrid model for $\Theta=\{A,B,C\}$.}
\label{Figure1}
\end{figure}

{\bf{Example for URR}}: Let's consider $\Theta=\{A,B,C\}$ with the DSm hybrid model as shown on the Figure \ref{Figure1}.
%
%
In this hybrid model $C\cap (A\cup B)=\emptyset$ (therefore $ A\cap C=\emptyset$ and $B\cap C=\emptyset$). We consider also the following two belief assignments 
$$m_1(A)=0.4\qquad m_1(B)=0.2\qquad m_1(A\cup B)=0.4$$
$$m_2(A)=0.2\qquad m_2(C)=0.3\qquad m_2(A\cup B)=0.5$$
\noindent
then the conjunctive operator provides for this DSm hybrid model a consensus on $A$, $B$, $C$, $A\cup B$, and $A\cap B$ with supporting masses
$$m_{12}(A)=0.36\quad m_{12}(B)=0.10\quad m_{12}(A\cup B)=0.20\quad m_{12}(A\cap B)=0.04\quad $$
\noindent
and partial conflicts between two sources on $A\cap C$ ,  $B\cap C$  and $C\cap (A\cup B)$ with
$$m_{12}(A\cap C)=0.12\quad m_{12}(B\cap C)=0.06\quad m_{12}(C\cap (A\cup B))=0.12$$
\noindent
Then with URR, the total conflicting mass $$m_{12}(A\cap C)+ m_{12}(B\cap C)+m_{12}(C\cap (A\cup B))=0.12+0.06 + 0.12=0.30$$ is uniformly (i.e. equally) redistributed to $A$, $B$, $C$ and $A\cup B$ because the sources support only these propositions. That is $n_{12}=4$ and thus $0.30/n_{12}=0.075$ is added to $m_{12}(A)$, $m_{12}(B)$, $m_{12}(C)$ and $m_{12}(A\cup B)$ with URR. One finally gets:

$$m_{12URR}(A)=m_{12}(A)+\frac{0.30}{n_{12}}=0.36+0.075=0.435$$
$$m_{12URR}(B)=m_{12}(B)+\frac{0.30}{n_{12}}=0.10+0.075=0.175$$
$$m_{12URR}(C)=m_{12}(C)+\frac{0.30}{n_{12}}=0.00+0.075=0.075$$
$$m_{12URR}(A\cup B)=m_{12}(A\cup B)+\frac{0.30}{n_{12}}=0.20+0.075=0.275$$

\noindent
while the others remain the same. That is $m_{12URR}(A\cap B)=0.04$. Of course, one has also
$$m_{12URR}(A\cap C)=m_{12URR}(B\cap C)=m_{12URR}(C\cap (A\cup B))=0$$

\noindent {\bf{Example for MURR}}:   Let's consider the same frame, same model and same bba as in previous example. In this case the total conflicting mass 0.30 is uniformly redistributed to the sets $A$, $B$, $A\cup B$, and $A\cap B$ only, i.e. to the sets whose masses, after applying the conjunctive rule to the given sources, are non-zero.
Thus $n_{12 }= 4$, and $0.30/4 = 0.075$. Hence: 

$$m_{12MURR}(A)=0.36+0.075 = 0.435$$
$$m_{12MURR}(B)=0.10+0.075=0.175$$
$$m_{12MURR}(A\cup B)=0.20+0.075=0.275$$
$$m_{12MURR}(A\cap B)=0.04+0.075=0.115$$

\section{Partially Uniform Redistribution Rule}

It is also possible to do a uniformly partial redistribution, i.e. to uniformly redistribute the conflicting mass only to the sets involved in the conflict. For example, if $m_{12}(A\cap B)=0.08$ and $A\cap B=\emptyset$, then 0.08 is equally redistributed to $A$ and $B$ only, supposing $A$ and $B$ are both non-empty, so 0.04 assigned to $A$ and 0.04 to $B$.\\

\noindent $\forall A\neq \emptyset$, one has the Partially Uniform Redistribution Rule (PURR) for two sources

\begin{equation}
m_{12PURR}(A)=m_{12}(A)+\frac{1}{2}\sum_{\substack{X_1,X_2\in G^\Theta \\ X_1\cap X_2=\emptyset \\ÊX_1=A\ \text{or} \ X_2=A}} m_1(X_1)m_2(X_2)
\label{eq:URR12}
\end{equation}

\noindent
where $m_{12}(A)$ is the result of the conjunctive rule applied to belief assignments $m_1(.)$ and $m_2(.)$. \\

\noindent For $s\geq 2$ sources to combine: $\forall A\neq \emptyset$, one has
\begin{equation}
m_{12\ldots sPURR}(A)=m_{12\ldots s}(A) +\frac{1}{s}
\sum_{\substack{X_1,X_2,\ldots ,X_s\in G^\Theta \\ X_1\cap X_2\cap \ldots \cap X_s=\emptyset \\
\text{at least one}\ X_j=A, j\in\{1,\ldots,s\}}} \text{Card}_A(\{X_1,\ldots,X_s\}) \prod_{i=1}^{s}m_i(X_i)
\label{eq:URR1s}
\end{equation}

\noindent
where $\text{Card}_A(\{X_1,\ldots,X_s\})$ is the number of $A$'s occurring in $\{X_1,X_2,\ldots,X_s\}$.\\

\noindent
If $A=\emptyset$, $m_{12PURR}(A)=0$ and $m_{12\ldots sPURR}(A)=0$.

\section{Example for PURR}

Let's take back the example of section \ref{sec3}. Based on PURR, $m_{12}(A\cap C)=0.12$ is redistributed as follows: 0.06 to $A$ and 0.06 to $C$; $m_{12}(B\cap C)=0.06$ is redistributed as follows: 0.03 to $B$ and 0.03 to $C$; and $m_{12}(C\cap (A\cup B))=0.12$ is redistributed in this way: 0.06 to $C$ and 0.06 to $A\cup B$. Therefore we finally get
$$m_{12PURR}(A)=m_{12}(A)+\frac{0.12}{2}=0.36+0.06=0.42$$
$$m_{12PURR}(B)=m_{12}(B)+\frac{0.06}{2}=0.10+0.03=0.13$$
$$m_{12PURR}(C)=m_{12}(C)+\frac{0.12}{2}+\frac{0.06}{2}+ \frac{0.12}{2}=0.15$$
$$m_{12PURR}(A\cup B) =m_{12}(A\cup B)+\frac{0.12}{2}=0.20+0.06=0.26$$

\noindent
while the others remain the same. That is $m_{12PURR}(A\cap B)=0.04$. Of course, one has also
$$m_{12PURR}(A\cap C)=m_{12PURR}(B\cap C)=m_{12PURR}(C\cap (A\cup B))=0$$

\section{Neutrality of vacuous belief assignment}

Both URR (with MURR included) and PURR are commutative and quasi-associative, and they verify the neutrality of Vacuous Belief Assignment (VBA): since any bba $m_1(.)$ combined with the VBA defined on any frame $\Theta=\{\theta_1,\ldots,\theta_n\}$ by $m_{VBA}(\theta_1\cup\ldots \cup \theta_n)=1$, using the conjunctive rule, gives $m_1(.)$, so no conflicting mass is needed to transfer.

\section{Conclusion}

Two new simple rules of combination have been presented in the framework of DSmT which have a lower complexity than PCR5. These rules are very easy to implement but from a theoretical point of view remain less precise in their transfer of conflicting beliefs since they do not take into account the proportional redistribution with respect to the mass of each set involved in the conflict. So we cannot reasonably expect that URR or PURR outperforms PCR5 but they may hopefully appear as good enough  in some specific fusion problems when the level of total conflict is not important. PURR does a more refined redistribution that URR and MURR but it requires a little more calculation.

\end{document}